\documentclass{article}
\usepackage[dblblindworkshop, final]{neurips_2025}

\usepackage{amsmath}
\usepackage{subcaption}
\usepackage{graphicx} 
\usepackage[utf8]{inputenc} 
\usepackage[T1]{fontenc}    
\usepackage{amsmath}
\usepackage[colorlinks=true, urlcolor=blue, citecolor=black]{hyperref}      
\usepackage{url}            
\usepackage{booktabs}       
\usepackage{amsfonts}       
\usepackage{nicefrac}       
\usepackage{microtype}      
\usepackage{xspace}
\usepackage{cleveref}
\usepackage[most]{tcolorbox}
\usepackage{listings}
\usepackage{bbold}
\usepackage{enumitem}
\usepackage{algorithm}
\usepackage{algpseudocode}
\usepackage{ifthen}
\usepackage{fontawesome5}
\usepackage{amsthm}
\usepackage{mathtools}
\usepackage{pifont}

\usepackage[varqu]{zi4}     
\usepackage{upquote}        

\graphicspath{{figures/}}

\newcommand{\study}{\textsc{Self-Study}\xspace}
\newcommand{\cartridge}{\textsc{Cartridge}\xspace}
\newcommand{\cartridges}{\textsc{Cartridges}\xspace}
\newcommand{\abcart}{\textsc{AblationCartridge}\xspace}
\newcommand{\verdict}{\textsc{Verdict}\xspace}

\newcommand{\longhealth}{\textsc{LongHealth}\xspace}
\newcommand{\finance}{\textsc{FinanceBench}\xspace}
\newcommand{\genconvo}{\textsc{GenConvo}\xspace}
\newcommand{\arxiv}{\textsc{Arxiv}\xspace}

\newcommand{\llama}{\textsc{Llama}\xspace}

\newcommand{\llamathreeb}{\textsc{Llama 3.2 3B}\xspace}
\newcommand{\llamaeightb}{\textsc{Llama 3.1 8B}\xspace}

\newcommand{\llamaoneb}{\textsc{Llama-1B}\xspace}
\newcommand{\qwen}{\textsc{Qwen3}\xspace}
\newcommand{\qwenthreefourb}{\textsc{Qwen3-4B}\xspace}

\newcommand{\eg}{\textit{e.g.}\xspace}

\newcommand{\etal}{\textit{et al.}\xspace}

\newcommand{\ctx}{\mathcal{C}}

\newcommand{\query}{q}

\newcommand{\vocab}{\mathcal{V}}
\newcommand{\llm}{\mathcal{F}}

\newtcolorbox{examplebox}[1][]{
  colback=lightgray!10,
  boxrule=0.75pt,
  title=#1,
  fonttitle=\bfseries,
  left=6pt,right=6pt,top=4pt,bottom=4pt,
  breakable,
  enhanced,
  fontupper=\small\ttfamily\raggedright,
  before upper=\setlength{\parskip}{3pt}\noindent
}

\usepackage[utf8]{inputenc} 
\usepackage[T1]{fontenc}    
\usepackage{hyperref}       
\usepackage{url}            
\usepackage{booktabs}       
\usepackage{amsfonts}       
\usepackage{nicefrac}       
\usepackage{microtype}      
\usepackage{xcolor}         

\title{Learned Structure in Cartridges: Keys as Shareable Routers in Self-Studied Representations}
\workshoptitle{Workshop on Mechanistic Interpretability}
\author{%
  Maurizio A. Diaz \\
  \texttt{mau@mit.edu} \\
}

\begin{document}

\maketitle

\begin{abstract}
A bottleneck for long-context LLM inference is the linearly growing KV cache. Recent work has proposed \cartridges, an approach which leverages offline compute to train a much smaller KV cache than is typically required for a full document (up to 40x less memory usage at inference time). In this paper, we present the first mechanistic exploration of the learned \cartridge key-value cache structure. In particular, we propose that (1) \cartridge keys act as stable, shareable retrieval routers for the compressed corpora and (2) most of the learned compression occurs within the \cartridge value vectors. We present empirical evidence of our routing theory across tasks, model families, and model sizes; for example, we can ablate the learned \cartridge key vectors between tasks with little performance loss. Finally, we propose a slight improvement in initialization called Sampled Chunk Initialization (SCI). We suggest that SCI can lead to faster \cartridge convergence than previously demonstrated in the literature. Our findings lay the groundwork for broader empirical study of \cartridge training optimization which may be crucial for further scaling.
\end{abstract}

\vspace{-2mm}
\section{Introduction}

As context windows grow for large language models (LLMs) \cite{vaswani2017attention}, users expect to process larger corpora. Common use-cases are large code bases \cite{nam2024using}, financial filings and market data \cite{islam2023financebench}, legislature \cite{guha2023legalbench}, or personal files \cite{arora2022can}. The most common solution is in-context learning (ICL) wherein we provide the full context to the downstream model; however, ICL-based inference relies on a key-value (KV) cache that grows linearly with context window utilization \cite{dong2022survey}. Because the cache grows linearly, we need to consider (1) the memory consumption of storing context for ICL and (2) the resulting reduction in throughput due to increased pressure on-device bandwidth and compute \cite{fu2024challenges}. This leaves users with an undesirable tradeoff between convenience and inference cost.

Recent work tackling long-context inference have primarily focused on server-level optimizations. Server-level approaches most notably include prompt (or prefix) caching, prefix-sharing (cascade and bifurcated attention, Hydragen \cite{juravsky2024hydragen}\cite{cascade-inference}\cite{athiwaratkun2024bifurcated}), and distributed approaches like ring attention.\cite{liu2023ring} However, these methods do not directly address compute and bandwidth pressures. In response, a growing roster of cache slimming protocols has emerged and we can divide them into two classes: token reducers and key-value compressors. Token reducers rely on summarization, chunking, or filtering to reduce corpora directly via heuristics or natural language \cite{jiang2023llmlingua}\cite{chuang2024learning}. Key-value compressors directly leverage sparsity or query oracles to project KV caches into lower rank spaces \cite{ge2023context}\cite{tang2024quest}.

While these approaches are promising, they typically come with a quality tradeoff. A more recent key-value compression method proposes a two-stage pipeline \cite{eyuboglu2025cartridges}:

\begin{enumerate}
    \item \textbf{Self-Study:} Exhaustively chunk the corpus and prompt a model to quiz itself on the content. These synthetic rollouts can be massively parallel and help build a diverse training dataset.
    \item \textbf{Context Distillation:} Once we have our conversational traces, we initialize the model with a fixed-size key-value cache of length $p \ll d_{\ctx}$ where $\ctx$ is our Corpus. Then, we can train on a context distillation objective to align our \cartridge-initialized model's next token distribution with the \study synthetic traces \cite{hinton2015distilling}\cite{snell2022learning}.
\end{enumerate}

\begin{figure*}[t]  
    \centering
    \includegraphics[width=\textwidth]{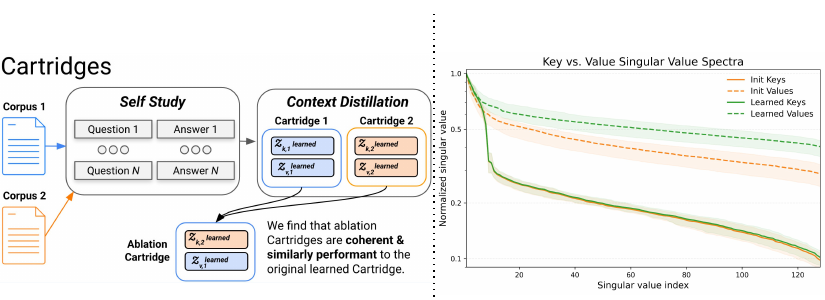}

    \caption{
        \textbf{(Left)} \cartridges learn to compress long-context documents by first generating synthetic conversations about the corpus and then training a small KV cache using the synthetic traces. This process is called \study. Leveraging a context distillation objective, we back-propagate \study traces into trainable KV caches while keeping the rest of the model frozen. \textbf{(Right)} Here we plot the layer-wise mean singular value spectra of a \llamaeightb \cartridge's KV vectors before and after training \cite{llama3}. The resulting key vectors are stable while the learned value vectors increase in singular value, representing a more efficient use of representation space due to compression.
    }
    \label{fig:headline}
\vspace{-3mm}
\end{figure*}

The synthetic trace generation process is \study and the final context-distilled cache is called a \cartridge. \cartridges exhibit interesting properties, including composability, high recall accuracy on benchmarks, and a tensor structure that is ideal for high throughput prefix-sharing inference engines \eg Tokasaurus \cite{juravsky2025tokasaurus}.

In this paper, we contribute a preliminary investigation into the mechanisms of a trained \cartridge. In particular, we study the structure of the learned \cartridge key and value vectors across model families, scale, and corpora. We can summarize our results as follows:

\begin{itemize}
    \item We observe that keys barely change during \cartridge training whereas values change significantly. We use singular value analysis to summarize this trend and argue that value vectors learn to use their representation dimensionality more effectively to maximize compression. This result likely explains why random initialization failed to converge in the original \cartridges paper.
    
    \item We ran an ablation experiment where we train two \cartridges with shared initializers on different tasks and then swap the learned key vectors but not the values. We note that responses remain coherent and the resulting performance loss is mild, which aligns with our first observation.
    
    \item We investigate a simple but effective initialization scheme for our \cartridge. Rather than using the first $p$ tokens of our corpus to initialize the \cartridge (the main method used in the original paper), we randomly sample chunks throughout the full corpus to maximize structural diversity. This approach leads to statistically significant improvements for training convergence.
\end{itemize}
\section{Preliminaries}

In this section, we position \cartridges as a new parameter efficient fine-tuning (PeFT) technique that borrows directly from prefix-tuning. We briefly summarize prior work on PeFT mechanistic interpretability, highlighting the lack of related work to prefix-tuning based PeFT methods. Next, we provide more details on \cartridges with a focus on the relevant experimental parameters and notation for our downstream experiments. Finally, we briefly cover the relevant datasets and benchmarks to reduce friction when framing our analyses in Sections 3 and 4.

\subsection{Related Work}

There is a significant body of work exploring task-specific LLM specialization. Common PeFT techniques are low-rank adapters (LoRA) and prefix-tuning, with the former being the most popular method \cite{hu2022lora}\cite{li2021prefix}. \cartridges primarily borrow from prefix-tuning: both methods prepend a learned representation to the input sequence and rely on the model to treat the trained prefix as real tokens. However, despite the widespread popularity of PeFT, there is comparatively little research focused on interpreting how LoRAs or prefix-tuning change models at a mechanistic level. 

\textbf{Interpreting LoRA} Recent interpretability work focuses on LoRA-driven mechanistic changes when tuning for narrowly scoped tasks. For example, Lee \etal studied sparse neural activations during LoRA for nuclear safety applications, while Nijasure et al. found that middle layers (5-15) and specific MLP projections drive ranking performance  \cite{lee2025mecha}\cite{nijasure2025relevance}. These works focus on neuron-level changes and layer contributions, whereas our analysis examines the geometric structure and transferability of learned representations.

\textbf{Interpreting Prefix-Tuning} Prefix-tuning, which is much less popular than LoRA in practice, has even less relevant interpretability work. However, Petrov \etal provide a theoretical analysis of when prefix-tuning works well \cite{petrov2024when}. They theoretically argue that prefix-tuning "cannot change the relative attention pattern over the content" and we find that this theory aligns strongly with our experimental findings. More concretely, they argue that key vectors have fixed directional biases established during pre-training; our ablation experiments (\Cref{fig:ablations}) support this claim directly on real downstream tasks.

\subsection{Relevant \cartridges Notation}

We choose to preserve notation from the original \cartridges paper to make cross-referencing easier for the reader. Taking that into consideration, we provide a brief overview of relevant definitions.

\textbf{LLMs} For an LLM $\llm$ we define the output distribution $p_{\text{model}}(\cdot) = \llm(\cdot | \mathbf{x})$ which is a categorical distribution over the model's vocabulary $\vocab$. Given a corpus $\ctx$ and queries $q$, we can define in-context learning (ICL) decoding of our queries as $\llm(\cdot | \ctx \oplus q)$.

\textbf{\cartridge} A \cartridge is a KV cache of size $p \ll n_\ctx$ which aims to augment an LLM $\llm$ such that the LLM behaves as if a Corpus $\ctx$ of length $n_\ctx$ were fully within context. Throughout the paper, we call this \cartridge $Z \in \mathbb{R}^{L \times p \times d_{\text{head}} \times 2}$ and define the augmented $\llm$ as $\llm_Z$. Given $l \in [1, L]$, where $l$ indexes our model layers, we can summarize the difference between traditional ICL and Cartridges as:

\begin{align*}
\hphantom{CARTRIDGE }\text{ICL KV Cache:} \quad
\underbrace{(\mathbf{k}[1], \mathbf{v}[1]), \ldots, (\mathbf{k}[n_\ctx], \mathbf{v}[n_\ctx])}_{\text{KV pairs for } \ctx}, \underbrace{(\mathbf{k}[n_\ctx + 1], \mathbf{v}[n_\ctx + 1]), \ldots}_{\text{KV pairs for } \query}
\end{align*}

\begin{align*}
\hspace{-3.0em}\text{\cartridge KV Cache:} \quad
\underbrace{(\mathbf{z}^{(l)}_{\text{k},1}, \mathbf{z}^{(l)}_{\text{v},1}), \ldots, (\mathbf{z}^{(l)}_{\text{k},p}, \mathbf{z}^{(l)}_{\text{v},p})}_{\text{Trainable pairs in } Z}, \underbrace{(\mathbf{k}[1], \mathbf{v}[1]), \ldots}_{\text{KV pairs for } \query}
\end{align*}

During \cartridge training, only the parameters composing our \cartridge $Z$ of sequence length $p$ are trainable while the rest of $\llm$ remains frozen.

\subsection{Datasets and Benchmarks}

For consistency, we re-use or reproduce the core datasets from the original \cartridges paper.

\textbf{\longhealth} is a question answering benchmark containing 20 fictional patients' detailed medical records. Each record is composed of multiple cases and each case is approximately 5,500 words long. For the \longhealth corpus $\ctx_\text{LongHealth}$, each prompt $q$ is a multiple choice question with 5 options. In accordance with the original paper, we use 10 full patient records concatenated to form a 100k+ token context corpus with 200 questions to measure accuracy. This formulation is different from the original \longhealth paper's evaluation setting which evaluates each patient independently. We prefer the \cartridges formulation because it showcases longer-context reasoning and requires that the model delineate between different patient cases.\footnote{In \longhealth, inter-patient differences can be subtle. For example, \texttt{patient\_01} might receive an ibuprofen prescription that's only 50mg different from \texttt{patient\_02}. By mixing the corpus, we require the model to perform multi-hop reasoning beyond memorizing just one ibuprofen dosage.}

\textbf{\genconvo} is a synthetic conversational dataset based on \finance's 2022 AMD 10-K corpus. Our \genconvo conversations are based on four different Q\&A prompts that encourage structural, multi-hop, mathematical, and factual retrieval reasoning (\Cref{app:genconvoprompts}). We reproduce \genconvo with Claude Sonnet 4.0 instead of 3.7 due to changes in rate limits between these models. To maximize throughput and balance API rate limit loads, we use inference-time scaling library \verdict for \genconvo synthesis \cite{kalra2025verdict}. Throughout the paper, we rely on \genconvo perplexity to measure a \cartridge's ability to compress financial reasoning. While the original \genconvo produces 16 answers per question, we opted to generate 50 manually validated answers per question for our evaluation set.

\textbf{\arxiv} is the tutorial \study dataset from the \cartridges GitHub. The \arxiv corpus is the source \texttt{.tex} file for the original \cartridges paper (40k tokens and 64k \study rollouts). We only use \arxiv to train \qwen's \abcart model used in \Cref{fig:ablations}. We do not run evaluations using \arxiv.





\section{What representation does a \textsc{Cartridge} learn?}

\begin{figure*}[t]  
    \centering
    \includegraphics[width=\textwidth]{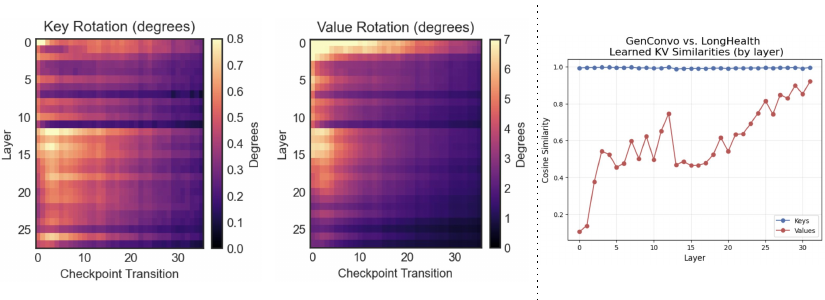}

    \caption{
        \textbf{(Left)} We reproduced a \longhealth \cartridge from the original paper. To do so, we trained a \llamathreeb \cartridge with length $p=2048$ for 3072 optimizer updates (batch\_size=64, sequence\_length=1024). We checkpointed the \cartridge every 96 optimizer steps and, for all layers $l \in [1, L]$, plotted the key and value vector rotation between each checkpoint. Notably, value rotations are often a full order of magnitude larger than key rotations and they continue late into the training process.
        \textbf{(Right)} We trained two \cartridges on separate tasks: \genconvo and \longhealth from the original paper. Afterwards, we plotted the layer-wise cosine similarity between the two fully trained \cartridges and note that their learned key vectors are highly similar. We explore this further in \Cref{fig:ablations} where we show that we can swap these key vectors with minimal downstream performance loss. On the other hand, the learned value vectors differ the most within the layers that experience the most vector rotations throughout learning.
    }
    \label{fig:kvs}
\vspace{-3mm}
\end{figure*}

In this section, we'll explore the evolution of the \cartridge key and value vectors throughout training. First, we'll analyze the geometric restructuring of the \cartridge KV cache from initialization to convergence (\Cref{fig:kvs}). Next, we'll formalize our singular value analysis and argue that it provides a summary of training dynamics. Finally, we'll run an ablation experiment to show that \cartridge key vectors can be swapped between learned tasks with minimal performance loss (\Cref{fig:ablations}).

\subsection{Observed Training Dynamics}

Let's briefly refresh the \study context distillation objective. First, we build a teacher distribution $p_{\text{teacher}}(\cdot) = \llm(\cdot | \ctx \oplus \query)$ over the synthetic conversations derived from the corpus we want to compress. In our case, we borrow the same conversation "seed prompts" from the original paper (\Cref{app:studyprompts}). We then train our student $p_{\text{student}}(\cdot) = \llm_Z(\cdot | \query)$ to minimize

\[
\mathcal{L} = \mathbb{E}_{(\ctx, \query) \sim \mathcal{D}_{\text{synth}}} \left[ \text{KL}(p_{\text{teacher}} \| p_{\text{student}}) \right]
\]

where $\mathcal{D}_{\text{synth}}$ is the synthetic conversation dataset from \study, and we optimize only the trainable parameters $Z = \{(\mathbf{z}^{(l)}_{\text{k},i}, \mathbf{z}^{(l)}_{\text{v},i})\}_{l=1}^L$ across all layers. For convenience, we say a \cartridge $Z$'s size is $p$.\footnote{The most intuitive analog for $p$ is to think of a \cartridge as a KV cache composed of $p$ tokens. Therefore, the context distillation objective is to compress the \study corpora into that KV cache such that $p$ is much smaller than the original corpus.} 

During this distillation, we note that our trainable key vectors rarely change whereas our value vectors change often; in particular, we measure these changes via the cosine similarity between \cartridge checkpoints (\Cref{fig:kvs}) \eg $\mathbf{cos}(Z^{(t)}, Z^{(t+1)})$. Our hypothesis is that the key vectors rotate minimally enough that given two \cartridges initialized via the same method, we could swap their key vectors arbitrarily. If this is true, it implies that \cartridge key vectors act as stable routers to compressed value payloads. The magnitude difference in rotations reflects this division of labor: while values require continuous refinement for compression, keys stabilize once effective routing is established.

\subsection{Singular Value Analysis}
\begin{algorithm}
\caption{Singular Value Analysis for Trained CARTRIDGE}
\label{alg:svd_analysis}
\begin{algorithmic}[1]
\State \textbf{Initialize:} $\mathcal{S} \leftarrow \{\}$ \Comment{Spectral collections}

\For{$\text{vector} \in \{\text{KEYS}, \text{VALUES}\}$}
    \State $\mathcal{S}[\text{vector}] \leftarrow \emptyset$
    \For{$l = 1$ \textbf{to} $L$} \Comment{Each layer}
        \State $T^{(l)} \gets$ \textsc{Extract}$(C, l, \text{vector})$ \Comment{$\mathbb{R}^{h \times p \times d_{\text{head}}}$}
        \State $X^{(l)} \gets$ \textsc{Reshape}$(T^{(l)}, [h \cdot p, d_{\text{head}}])$ \Comment{Flatten heads}
        \State $\sigma^{(l)} \gets$ \textsc{SVD\_Vals}$(X^{(l)})$ \Comment{Singular values}
        \State $\tilde{\sigma}^{(l)} \gets \sigma^{(l)} / \sigma_1^{(l)}$ \Comment{Normalize by largest}
        \State $\mathcal{S}[\text{vector}] \gets \mathcal{S}[\text{vector}] \cup \{\tilde{\sigma}^{(l)}[:k]\ $ \Comment{Fetch the top $k$ values}
    \EndFor
\EndFor
\State \textbf{return} \{\textsc{Median}$(\mathcal{S}[\text{KEYS}])$, \textsc{Median}$(\mathcal{S}[\text{VALUES}])$\}
\label{fig:sva}
\end{algorithmic}
\end{algorithm}

\begin{figure}[t]  
    \centering
    \includegraphics[width=\textwidth]{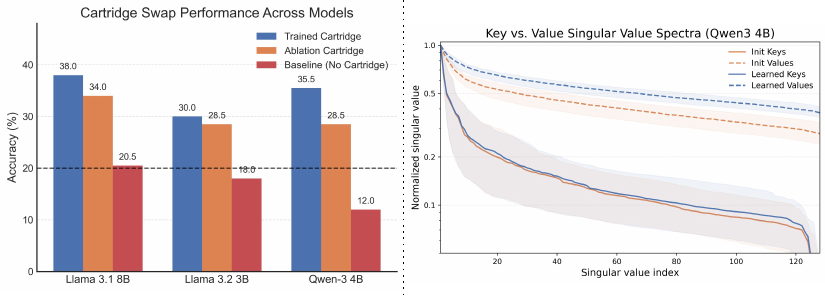}

    \caption{
        \textbf{(Left)} We present three \longhealth evaluation settings: a baseline with no \cartridge (red), a model with a \longhealth-trained \cartridge (blue), and a model where we swap its \longhealth-trained \cartridge key vectors with keys from a different task (orange). We call the latter an \textbf{\abcart.} For the \llama models, we swap in key vectors trained on \genconvo and for \qwen we use key vectors trained on \arxiv data \cite{qwen3}. While the vector swap leads to a slight performance loss, the \abcart still outperforms both a random choice baseline and the model's baseline performance.
        \textbf{(Right)} We reran our KV cache singular value analysis on \qwen. First, we noticed that \qwen exhibits the same training-time value vector singular increase of the \llama family. Second, \qwen's key vector singular values are higher variance than \llama which might explain the larger performance loss during ablation.
    }
    \label{fig:ablations}
\vspace{-3mm}
\end{figure}

While informative, we found the cosine similarity heatmaps too busy to establish a scalable, cross-model trend. As a response, we designed an analysis which runs a layer-wise singular value decomposition and then normalizes those values (\Cref{alg:svd_analysis}). In addition to reporting the median trend line, we also provide inter-quartile range bands (IQR) at the 25th and 75th percentile. Here's what we observed:

\textbf{Trends Across Scale} We found the high-level trend of stable routing keys and compressed value payloads consistent across model sizes (\Cref{app:svd}). It seems that the model's pretrained prior is strong enough that it's more effective to focus on value re-alignment and compression than finding a more effective key structure (\Cref{fig:ablations}).

\textbf{Differences Between Model Families} In \Cref{fig:ablations}, we leverage our singular value analysis to explore why \qwen responds differently to our ablation than \llama (-7\% vs -4-5\% performance drop). We observe that \qwen key vectors are more heterogeneous across layers than \llama (via the IQR bands). The wider IQR bands directly correspond to \qwen's larger performance drop, suggesting layer-wise key specialization reduces transferability. 

Furthermore, based on how little \cartridge key vectors drift throughout training we recognize that initialization might play a larger role in \study-based distillation than we thought. We explore this more in (\Cref{fig:convergence}) where we show statistically significant faster convergence by randomly sampling the corpus to initialize our \cartridge.

\begin{table}[b]
\centering
\caption{We ran a hypergeometric statistical test ($N=200$) on the Q\&A correctness overlap between trained and ablated \cartridges. We confirm that the overlap is statistically significant compared to random chance in 5-question multiple choice Q\&A, where $n_{\text{train}}$ and $n_{\text{ablated}}$ represent correct answers out of 200 questions and $n_{\text{overlap}}$ counts questions answered correctly by both. Notably, despite experiencing the largest performance loss due to ablation, \qwenthreefourb has the most overlap of the three models we tested.} 
\begin{tabular}{lccccr}
\toprule
Model & $n_{\text{train}}$ & $n_{\text{ablated}}$ & $n_{\text{overlap}}$& $p$-value \\
\midrule
Llama 3.1 8B & 76 & 68 & 34 & 0.0156$^*$ \\
Llama 3.2 3B & 60 & 57 & 28 & 0.0003$^*$ \\
Qwen-3 4B & 71 & 57 & 40 & $<$0.0001$^*$ \\
\bottomrule
\end{tabular}
\label{tab:overlap_significance}
\end{table}

\subsection{Key Vector Ablations}

Informed by our \cartridge KV cache singular value analysis, we decided to run an ablation experiment on the \longhealth benchmark used in the original paper (\Cref{fig:ablations}). For two trained \cartridges $Z_A = \{(\mathbf{z}^A_{\text{k},i}, \mathbf{z}^A_{\text{v},i})\}_{i=1}^p$ and $Z_B = \{(\mathbf{z}^B_{\text{k},i}, \mathbf{z}^B_{\text{v},i})\}_{i=1}^p$ from different tasks $\{A, B\}$, we construct the \abcart:

\[
Z_{\text{AB}} = \{(\mathbf{z}^B_{\text{k},i}, \mathbf{z}^A_{\text{v},i})\}_{i=1}^p
\]

In comparison to the \cartridge-enabled LLMs $\llm(Z_A)$ and $\llm(Z_\text{AB})$, we also define a baseline model $\llm(Z_\emptyset)$ which has no \cartridge. We call $\mathbf{z}^A_{\text{k},i}$ and $\mathbf{z}^B_{\text{k},i}$ \textbf{transferable} if $\text{ACC}(\llm(Z_\text{AB})) > \text{ACC}(\llm(Z_\emptyset))$ and the answer overlap between $\llm(Z_A)$ and $\llm(Z_{AB})$ is statistically significant (see: \Cref{tab:overlap_significance}).

For our ablation experiments, we train all models for 512 optimizer steps with a batch size of 128 and a packed sequence length of 1024. For all experiments, we train $Z_A$ on \longhealth and $Z_B$ on either \genconvo or \arxiv. The \cartridge sequence length $p$ is held consistent at 2048 and the initializing document is the default initializer from the \cartridges repository (\textbf{gradients.txt}, the Wikipedia article for gradients). Surprisingly, we find that keys from \textbf{First $k$ Token Initialization} are transferable for \llamathreeb, \llamaeightb, and \qwen. 
\section{The importance of good initialization}

In this section, we propose and briefly benchmark an improved initialization scheme for \cartridges. Inspired by our observation of training dynamics and KV cache geometry, we suggest that \cartridge initialization benefits from the structural diversity of random sampling.

\begin{figure*}[t]  
    \centering
    \includegraphics[width=\textwidth]{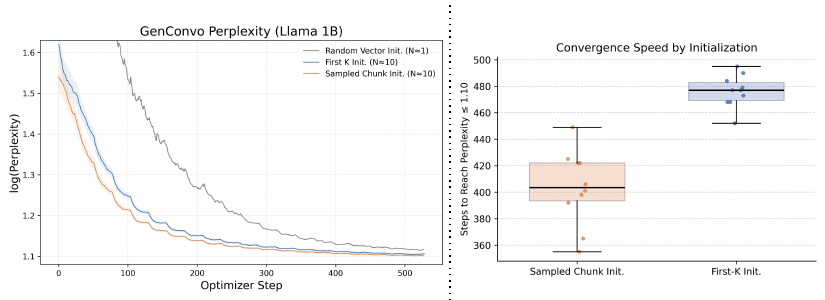}

    \caption{
        \textbf{(Left)} We plot the perplexity for 10 \llamaoneb \genconvo training runs with both Sampled Chunk Initialization (SCI) and the original paper's First $k$ Tokens Initialization. Additionally, we include a random vector initialization run for comparison. For all our SCI experiments, we chose chunksize=64 because it was the midpoint in our $n$-gram diversity vs. context length analysis (\Cref{app:ngram}).
        \textbf{(Right)} A different view on the perplexity graph, we can visualize convergence speed over our runs as a box plot. Setting a target threshold of $\text{perplexity}=1.10$ to define convergence, we can run a paired $t$-test to confirm that SCI converges at a statistically faster rate ($p<0.05$) than the original paper's \textbf{First-$k$ Token Initialization} scheme.
        }
    \label{fig:convergence}
\vspace{-3mm}
\end{figure*}

\subsection{Known Initialization Schemes}

The original \cartridges paper explored a few approaches for \cartridge initialization:

\begin{itemize}
    \item \textbf{Random Vector Initialization}: random vector initialization (RVI) for prefix tuning is known to be unstable from prior literature.\cite{li2021prefix} In the original \cartridges paper, the authors note that random vector initialization fails to converge and performs poorly compared to other methods. In hindsight, it's clear why: instead of focusing on value compression, a randomly initialized prefix is burdened with a joint optimization problem of both routing \textit{and} compression.
    \item \textbf{First-$k$ Token Initialization}: using the first $k$ tokens from the target corpus $C$ is a natural first-pass approach and is the initialization scheme used by the original paper. By sampling from real text, we are leveraging structures the model is already comfortable working with.
    \item \textbf{Summary-Based Initialization}: alternatively, we could use a lighter-weight summarization model to initialize a \cartridge. While attractive, we de-prioritized this approach to avoid explicitly stacking models which can introduce further uncertainty when benchmarking performance.
\end{itemize}

\subsection{Sampled Chunk Initialization}

\begin{algorithm}[h]
\caption{Sampled Chunk Initialization (SCI)}
\label{alg:sci}
\begin{algorithmic}[1]
\Require Corpus $\ctx$ with $n_{\text{total}}$ tokens
\Require Target cache size $p$, chunk size $c$

\State $\mathbf{x} \gets$ \textsc{Tokenize}$(\ctx)$ \Comment{Full corpus tokenization}
\State $n_{\text{chunks}} \gets \lfloor p / c \rfloor$

\State $\mathbf{s} \sim \text{Uniform}(\{0, 1, \ldots, n_{\text{total}} - c\}, n_{\text{chunks}})$

\State $\mathbf{x}_{\text{init}} \gets []$
\For{$i = 1$ \textbf{to} $n_{\text{chunks}}$}
    \State $\mathbf{chunk}_i \gets \mathbf{x}[s_i : s_i + c]$
    \State $\mathbf{x}_{\text{init}} \gets \mathbf{x}_{\text{init}} \oplus \mathbf{chunk}_i$
\EndFor

\State $\mathbf{x}_{\text{init}} \gets \mathbf{x}_{\text{init}}[:p]$ \Comment{Truncate to target size}

\State \textbf{return} \textsc{ForwardPass}$(\llm, \mathbf{x}_{\text{init}})$
\end{algorithmic}
\end{algorithm}

We present Sampled Chunk Initialization (SCI) (\Cref{alg:sci}), a principled alternative to the original paper's \textbf{First-$k$ Token Initialization}. In \Cref{fig:convergence} we show that SCI converges faster than both \textbf{First-$k$ Token Initialization} and random initialization. In (\Cref{app:svd_inits}), we present further singular value analyses of the three initialization methods in (\Cref{fig:convergence}). We note that our trained \cartridges using RVI hardly deviate from their spectra at initialization-time. Notably, the initialization spectra stays close to 1, aligning with random matrix theory which observes that randomly sampled orthogonal vectors maintain singular values close to one \cite{Tao2012}. This differs significantly from the spectra from token-based initialization methods which decay in a consistent, structured manner.

To confirm that SCI is better than \textbf{First-$k$ Token Initialization}, we run a simple paired-$t$ statistical test. Let $\mu$ represent the mean steps to reach perplexity threshold 1.10. Then if we can reject the null hypothesis $\{H_0: \mu_{\text{SCI}} = \mu_{\text{first-k}}\}$ against the alternative $\{H_1: \mu_{\text{SCI}} < \mu_{\text{first-k}}\}$ we can claim statistically significant better convergence. In \Cref{fig:convergence} we reject $H_0$ with $p < 0.05$, confirming faster convergence.
\section{Limitations and Future Work}

\subsection{Limitations}

\textbf{Scale} We ran all experiments on single A10G, A100, and H100 spot instances. Future work should explore larger models and train for longer periods of time. Except for our pure 1:1 reproduction of the original paper in \Cref{fig:kvs}, the rest of our models are approximately 4-6x optimizer steps under-trained compared to the best performing models from the original paper. In the appendix, we show that our singular value analysis holds for the 1:1 reproduced model (\Cref{app:svd_repro}).

\textbf{Baselines} When testing initialization schemes, we had to decide between method diversity and statistical significance. Future work could explore more initialization techniques, including more robust baselines \eg summary-based initializations. Additionally, it would be insightful to test convergence on multiple tasks instead of just \genconvo.

\subsection{Future Work}

\textbf{Training and Serving} Given that keys change little throughout training, it might be worth training \cartridges with fully frozen key vectors. Beyond being an easy win for training efficiency, there might be desirable properties of frozen keys at inference-time. Suppose we only need trainable values for corpora compression: can we imagine a serving engine that exploits this to hot-swap \cartridge value vectors at inference time while maintaining fixed key vectors? This may be a promising direction given the rise of prefix-sharing optimized serving engines like Tokasaurus \cite{juravsky2025tokasaurus}.

Further research may also explore if there exists key vector initialization structures that are particularly good for certain tasks; or, even better, key vector initializations that are universally well-performing.

\textbf{Implications for Prefix-Tuning} While we focused on \cartridges, mechanistic interpretability work for prefix-tuning remains sparse. Our findings provide a strong experimental foundation to support Petrov \etal's theoretical arguments regarding the limitations of prefix-tuning. Further mechanistic work could explore the overlap between \cartridge training dynamics and prefix-tuning; for example, do prefix-tuned key vectors also exhibit stability? Or, can they learn significantly novel attention patterns? 

Beyond supporting existing theory, our work raises questions about which tasks are appropriate for \cartridges. Future work could explore counter-examples to the patterns we observed during our experiments. Does there exist some task where keys must significantly reroute to perform well and does this delay convergence? Counter-examples could help formalize a boundary between compression-friendly \cartridge tasks and tasks better suited for prefix-tuning or LoRA.

\section{Conclusion}

In this paper, we present the first mechanistic study of \cartridges, a prefix-tuning based method for compressing long-context corpora prior to inference. Our analyses showed that \cartridge key vectors act as a stable routing mechanism throughout training whereas the value vectors absorb most of the representational load required for compression. Based on this observation, we ran a set of key vector ablation experiments that confirmed that \cartridge key vectors are shareable across tasks. Building on our mechanistic study of key vector structure, we introduce a simple sampled-chunk initialization scheme (SCI) and showed that it accelerates convergence.

Collectively, these findings provide experimental validation of Petrov \etal's theoretical constraints on prefix-tuning, suggesting that there is a fundamental division of labor between keys and value vectors during prefix-tuning. Rather than being limitations, we suggest that these constraints could inform further training- and inference-time optimizations for \cartridges and other prefix-tuning methods.

By presenting an empirical study of \cartridges' learned structure, we hope to invite further mechanistic research and discussion for understanding PeFT methods more broadly.

\begin{ack}
We would like to thank Prime Intellect for providing the compute for this project, Ileana Rugina for feedback, and Sabri Eyuboglu for open-sourcing the code that made this project possible.
\end{ack}

\bibliographystyle{plain}
\bibliography{references}

\newpage
\appendix

\section{N-Gram Diversity of Initialization Methods}
\label{app:ngram}

\begin{figure*}[h]  
    \centering
    \includegraphics[width=\textwidth]{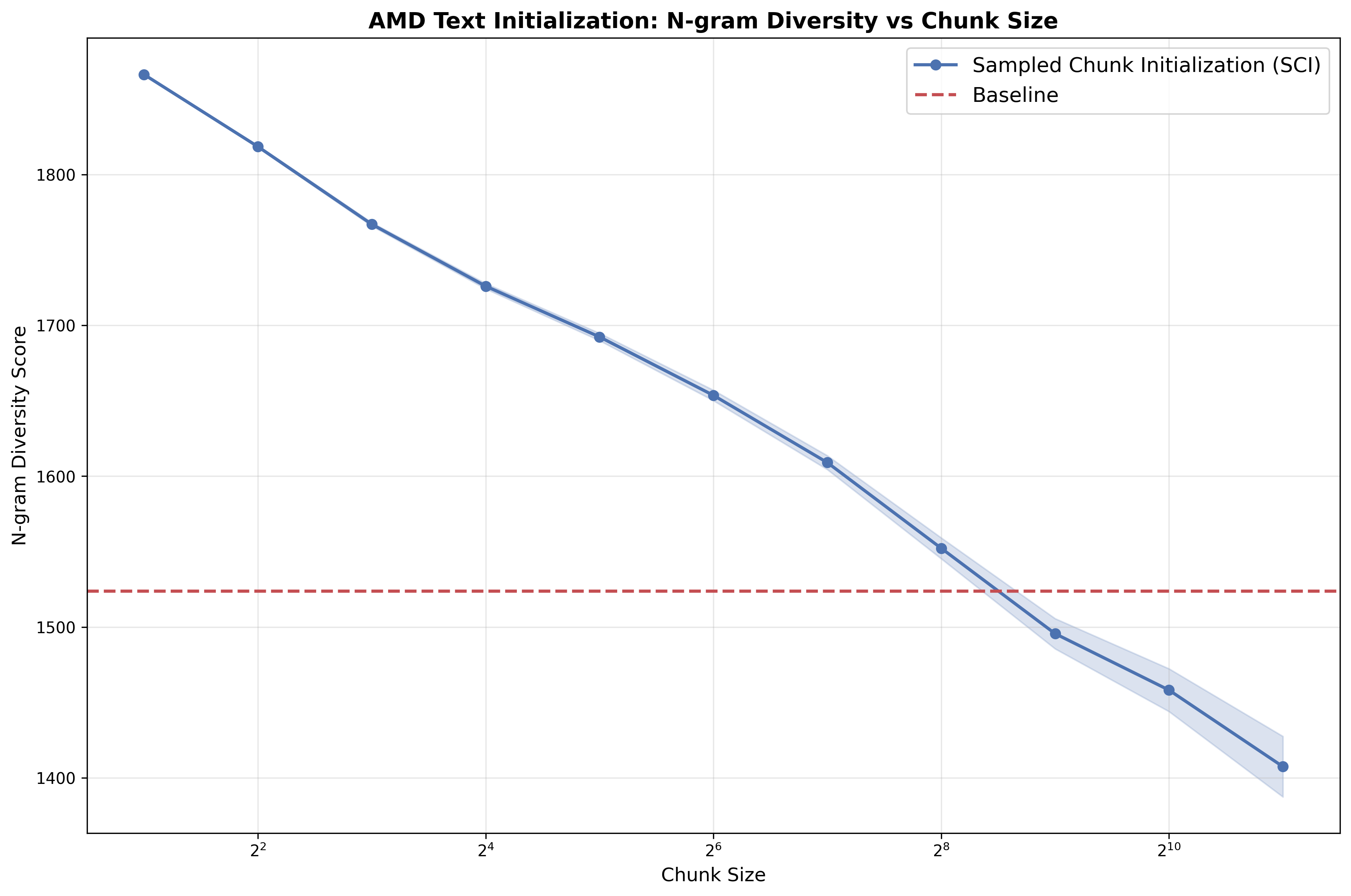}

    \caption{
    Here we plot the N-gram diversity of Sampled Chunk Initialization (SCI) vs. the \textbf{First $k$ Token Initialization} baseline. We note that $2^6$ is approximately the midway point when trading diversity for chunk length, so we chose 64 as the chunksize when running our experiments in \Cref{fig:convergence}.
    }
    \label{fig:ngram}
\vspace{-3mm}
\end{figure*}

\section{Additional Singular Value Analyses}
\label{app:svd}

Here we present additional supporting singular value analysis prompts to support our observations across scale, model families, and training regimes.

\subsection{\llama Across Scales}
\label{app:svd_llama}

\begin{figure*}[h]  
    \centering
    \includegraphics[width=\textwidth]{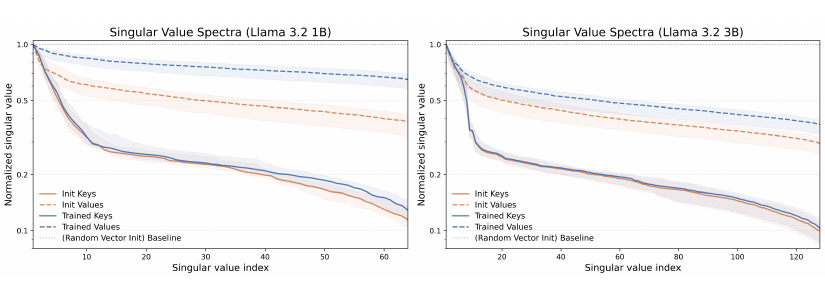}

    \caption{
    Here we have the singular value spectra for both \llama 1B and \llama 3B. We observe the same trends from the main paper: keys remain stable and values generally shift up in singular value. Both these models were trained on \longhealth with the paper's \textbf{First $k$ Token Initialization} scheme. However, they are relatively undertrained at only 512 optimizer updates with a batch size of 64 and packed sequence length of 1024. In \Cref{app:svd_repro}, we note that training further reduces the IQR band variance for \llama 3.2 3B.
    }
    \label{fig:llama_spectra}
\vspace{-3mm}
\end{figure*}

\subsection{\qwen Across Scales}
\label{app:svd_qwen}

\begin{figure*}[h]  
    \centering
    \includegraphics[width=\textwidth]{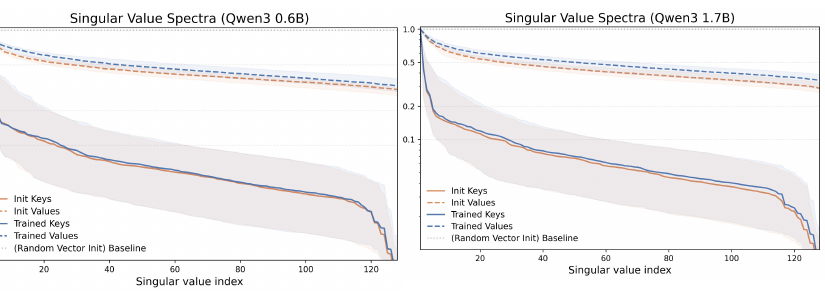}

    \caption{
    We present the singular value spectra for \qwen 0.6B and \qwen 1.7B on \longhealth. Notably, we see the higher IQR badnd variance from \Cref{fig:ablations} repeated for the other \qwen models. Considering the results from \llama, it's possible that further training could reduce this variance given that these models are under-trained at only around 320 optimizer steps. Despite this undertraining, we still see a general upward singular value trend for the \cartridge value vectors; however, we cannot claim statistical significance for the smaller \qwen models without more data.
    }
    \label{fig:qwen_spectra}
\vspace{-3mm}
\end{figure*}

\subsection{\llama Strict Replication}
\label{app:svd_repro}

\begin{figure*}[h]  
    \centering
    \includegraphics[width=\textwidth]{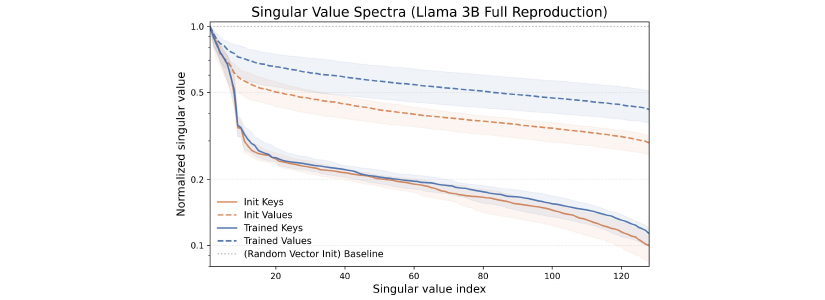}

    \caption{
    This is the singular value spectra of a \llama 3B model trained for 3072 steps on \longhealth. We see the trend holds at scale; notably, given more optimizer steps the value vectors rose in singular value more than the under-trained \llama 3B from \Cref{app:svd_llama}.
    }
    \label{fig:llama_full}
\vspace{-3mm}
\end{figure*}

\subsection{Across Different Initializations}
\label{app:svd_inits}

\begin{figure*}[h]  
    \centering
    \includegraphics[width=\textwidth]{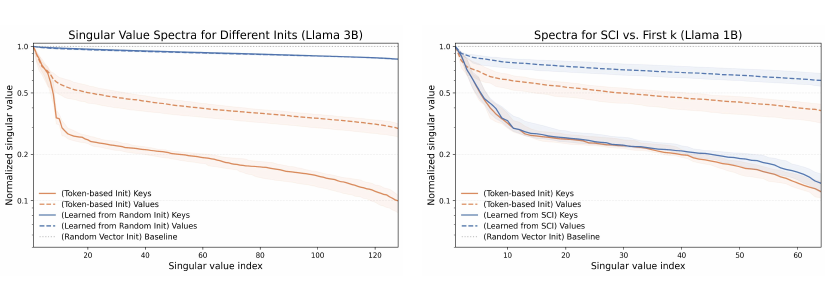}

    \caption{
    We compare two initialization methods against \textbf{First $k$ Tokens Initialization} \textbf{(Left)} Here the singular value spectra of random vector initialization (RVI) is remarkably different from our token-based approaches. Notably, both RVI keys and vectors stay close to the random orthogonal vector baseline. This suggests that keys fail to learn an effective routing structure for downstream tasks.
    \textbf{(Right)} Here is the singular value spectra for SCI. It looks more like what we expect to see from a successfully trained \cartridge: stable keys and shifted values due to compression.
    }
    \label{fig:inits}
\vspace{-3mm}
\end{figure*}

\newpage
\section{Prompts}
\label{app:prompts}

\subsection{\study Prompts}
\label{app:studyprompts}

We use the same \study prompts from the \href{https://github.com/HazyResearch/cartridges/tree/main/cartridges}{Cartridges GitHub}.

\lstdefinestyle{pyplain}{
  language=Python,
  basicstyle=\ttfamily\small,
  numbers=left,
  numberstyle=\tiny,
  stepnumber=1,
  numbersep=8pt,
  frame=single,
  framerule=0.4pt,
  breaklines=true,
  breakatwhitespace=false,
  showstringspaces=false,
  showtabs=false,
  showspaces=false,
  keepspaces=true,
  tabsize=4,
  columns=fullflexible,
  upquote=true,
  keywordstyle=\bfseries,
  commentstyle=\itshape,
  stringstyle=\itshape,
}

\begin{lstlisting}[style=pyplain]
def structuring_seed_prompt(**kwargs):
    DATA_FORMATS = [
        "JSON",
        "YAML",
        "TOML",
        "INI",
        "XML",
        "plain text",
    ]

    data_format = random.choice(DATA_FORMATS)

    EXAMPLES = [
        (
            "Can you structure the information in {{subsection}} of {{document}} related to {{something specific}} "
            f"in the following format: {data_format}? "
            "Be sure to include precise information like any dates, times, names, and numerical values.'"
        ),
        (
            "Can you structure the information in {{subsection}} of {{document}} "
            f"in the following format: {data_format}? "
            "Be sure to include precise information like any dates, times, names, and numerical values.'"
        ),
    ]

    example = random.choice(EXAMPLES)

    return (
        f"Please generate a single chat message instructing an LLM to structure the information in {data_format}. "
        "Output only the chat message itself and absolutely nothing else. "
        "Make sure it is clear what section and document you are asking about. "
        f"The message can follow the following template, filling in details from the corpus: \n\n'{example}'"
    )


def summarization_seed_prompt(**kwargs):
    prompts = [
        (
            "Please generate a single chat message instructing an LLM to summarize part of the corpus. "
            "Make sure the instruction is very explicit about the section of the corpus that you want to summarize. "
            "Include details (ids, names, titles, dates, etc.) that make it clear what you are asking about. "
        ),
        (
            "Please generate a single chat message instructing an LLM to summarize a section. "
            "Make sure the instruction is explicit about the section that should be summarized and the document it is from."
        ),
    ]
    prompt = random.choice(prompts)
    return prompt


def question_seed_prompt(**kwargs):
    prompts = [
        (
            "Generate a question for an LLM that will test its knowledge of the information in the corpus above. "
            "In your question be sure to include details (ids, names, titles, dates, etc.) that make it clear what you are asking about. "
            "Output only a single question. Do NOT include any other text or explanation other than the question."
        ),
        (
            "Generate a message for an LLM that will test its knowledge of the information in the corpus above."
            "Be sure to include details (ids, names, titles, dates, etc.) in the question so that it can be answered without access to the corpus (i.e. closed-book setting). "
            "Output only a single question. Do NOT include any other text or explanation other than the question."
        ),
        (
            "You are helping to quiz a user about the information in the corpus. "
            "Please generate a question about the subsection of the corpus above. "
            "Be sure to include details (ids, names, titles, dates, etc.) in the question to make it clear what you are asking about. "
            "Answer only with the question, do not include any other text."
        ),
    ]
    prompt = random.choice(prompts)
    return prompt


def use_case_seed_prompt(**kwargs):
    prompt = (
        "You are working to train a language model on the information in the following corpus. "
        "Your primary goal is to think about practical, real-world tasks or applications that someone could achieve using the knowledge contained within this corpus. "
        "Consider how a user might want to apply this information, not just recall it. "
        "After considering potential use cases, your task will be to generate a sample question that reflects one of these downstream applications. "
        "This question/instruction/task should be something a user, who has access to this corpus, might ask when trying to accomplish their specific goal. "
        "Output only a single question. Do NOT include any other text or explanation other than the question."
    )
    return prompt


def creative_seed_prompt(**kwargs):
    prompt = [
        (
            "You are having a creative conversation inspired by the information in the corpus. "
            "Please generate a question for your conversation partner to start off the discussion. "
            "Answer only with the question, do not include any other text."
        ),
    ]
    return random.choice(prompt)


def generic_seed_prompt(**kwargs):
    return (
        f"Please generate a single chat message to begin a conversation about the information in the corpus. Ask a question about the corpus or make a request."
    )
\end{lstlisting}

\subsection{\genconvo Prompts}
\label{app:genconvoprompts}
We use modifications of the original \genconvo prompts from the \cartridges paper. These prompts yield similar Q\&A pairs but encourage better instruction following from the generating model.

\begin{examplebox}[Factual Prompt Template]
Generate a factual recall question about a specific entity, date, or name from the document.

\textbf{Format:} "Who/What/When [specific question]?"  

\textbf{Answer:} Must be an exact entity name, date, or proper noun from the document (2–4 words max).

The answer should be unambiguous and directly stated in the document.
\end{examplebox}

\begin{examplebox}[Reasoning Prompt Template]
Generate a mathematical reasoning question requiring calculation over document values.

\textbf{Format:} "What is the [percentage/ratio/difference] of [specific calculation]?"  

\textbf{Answer:} Must be a precise number with units (e.g., "12.5\%", "\$2.3M", "1.8x").

Question should require combining 2+ values from different parts of the document.
\end{examplebox}

\begin{examplebox}[Counting Prompt Template]
Generate a counting question about document structure or content frequency.

\textbf{Format:} "How many [items] are [condition]?"  

\textbf{Answer:} Must be a single integer (e.g., "7", "23").

Focus on countable elements like sections, tables, mentions of specific terms, or occurrences.
\end{examplebox}

\begin{examplebox}[Synthesis Prompt Template]
Generate a multiple choice question testing document comprehension.

\textbf{Format:} Question with 5 options (A/B/C/D/E).  

\textbf{Answer:} Single letter (A, B, C, D, or E). E always means "There is not enough information to answer the question".

Question should require understanding main themes, risks, or business model elements.
\end{examplebox}
\end{document}